# Embedding-Driven Diversity Sampling to Improve Few-Shot Synthetic Data Generation

Ivan Lopez[1,2,†], Fateme Nateghi Haredasht[3], Kaitlin Caoili[4],
Jonathan H Chen[2,5,6,7], Akshay Chaudhari[8,9]

[1]Stanford University School of Medicine, Stanford, CA, USA.;
[2]Department of Biomedical Data Science, Stanford, CA, USA.
[3]Center for Biomedical Informatics Research, Stanford, CA, USA.
[4]The Ohio State University College of Medicine, Columbus, OH, USA
[5]Division of Hospital Medicine, Stanford University School of Medicine, Stanford, CA, USA.
[6]Clinical Excellence Research Center, Stanford School of Medicine, Stanford, CA, USA.
[7]Department of Medicine, Stanford, CA, USA.
[8]Department of Radiology, Stanford, CA, USA.
[9]Cardiovascular Institute, Stanford, CA, USA.

January 10, 2025

†Corresponding Author: ivlopez@stanford.edu

## Abstract

Accurate classification of clinical text often requires fine-tuning pre-trained language models, a process that is costly and time-consuming due to the need for high-quality data and expert annotators. Synthetic data generation offers an alternative, though pre-trained models may not capture the syntactic diversity of clinical notes. We propose an embedding-driven approach that uses diversity sampling from a small set of real clinical notes to guide large language models in few-shot prompting, generating synthetic text that better reflects clinical syntax. We evaluated this method using the CheXpert dataset on a classification task, comparing it to random few-shot and zero-shot approaches. Using cosine similarity and a Turing test, our approach produced synthetic notes that more closely align with real clinical text. Our pipeline reduced the data needed to reach the 0.85 AUC cutoff by 40% for AUROC and 30% for AUPRC, while augmenting models with synthetic data improved AUROC by 57% and AUPRC by 68%. Additionally, our synthetic data was 0.9 times as effective as real data, a 60% improvement in value.

# 1 Introduction

Free-text notes in electronic health records (EHRs) contain valuable information not captured in structured fields, such as symptoms, diagnoses, disease progression, social determinants of health, and patient perspectives[1,2]. Annotating this data can enable numerous important applications in research and quality improvement, including cohort selection[3], electronic phenotyping[4–6], and predictive modeling[7–12]. Despite the wealth of valuable information in EHRs, automated annotation of clinical concepts remains a significant challenge[13,14], often requiring manual annotation for high-quality label extraction. Manually annotating such data for healthcare tasks is costly, labor-intensive, and challenging to scale to large datasets[15], which limits the availability of large, annotated datasets for clinical natural language processing (NLP) tasks. This lack of labeled data significantly impedes the progress of developing accurate NLP tools for supervised healthcare applications.

Synthetic data presents a promising solution to this issue, which provides several benefits. Namely, synthetic data can be generated automatically and be designed to closely replicate real-world patient records, providing a scalable and efficient alternative to manual data annotation[16,17]. Prior approaches have explored generating synthetic clinical data using statistical models such as generative adversarial networks and variational autoencoders[18,19]. However, a gap persists in effectively leveraging these tools to generate synthetic text that aligns with the syntactic diversity and quality of real-world clinical notes[20].

Recently, large language models (LLMs) have demonstrated impressive performance in natural language understanding tasks, such as identifying patient conditions[21], summarizing patient histories[22], and generating conversational responses[23]. Some approaches have been developed to generate synthetic clinical data using LLMs. One prominent example is GatorTronGPT[24], a generative model designed to synthesize clinical text by leveraging both clinical and general English corpora for model pretraining. While GatorTronGPT has shown promising results in generating coherent and contextually appropriate clinical text, it presents some limitations, such as requiring significant computational resources to develop, which is beyond the reach of most healthcare institutions. Additionally, GatorTronGPT's performance, although strong within a single-institution setting in which it was trained, may not generalize well when deployed across different healthcare systems that considerably varied syntactic diversity in clinical text[25]. This lack of generalizability restricts its broad application, as models trained in one institution may perform suboptimally in others due to differences in the structure and terminology of clinical notes as well as the type of clinical practice.

Beyond GatorTronGPT, other approaches have attempted to use LLMs to generate synthetic clinical data. Li et al.[26] explored two directions for clinical data generation with LLMs: data-to-label and label-to-data. Their study demonstrated that while LLMs can be effective in generating synthetic clinical data, they often fail to replicate the diversity and nuance of real-world clinical data[13]. Another limitation of LLMs in this context is their tendency to overfit to the specific terminologies and patterns present in the training data, leading to repetitive and less varied synthetic text. Further efforts have been made by Ive et al.[27], who generated artificial mental health records for NLP tasks. While their approach introduced diversity into the generated data, it still did not fully address the complexity of medical language, particularly when it comes to capturing the varied synonyms, abbreviations, negations, and idiomatic expressions commonly found in real-world clinical notes.

Overall, while some progress has been made in generating synthetic clinical data using LLMs, there is a large need to develop frameworks that generate synthetic data with representative syntactic diversity and effective utility on downstream tasks. In this paper, we devise an embedding-driven approach that leverages diversity sampling of a small amount of real-world clinical notes for few-shot prompting to generate a synthetic dataset. We depict that this synthetic dataset improves the efficiency of model fine-tuning tasks and generates data points that more closely align with the syntactic style of clinical notes[28]. We refer to this approach as 'diversity sampling' across the remainder of the paper. Our work presents three key contributions. First, we validate the authenticity of our embedding-driven diversity sampling pipeline. Performing Uniform Manifold Approximation and Projection (UMAP)[29] on contextualized sentence embeddings, we demonstrate that the average cosine similarity between synthetic data and real-world data is higher than that of zero-shot generated synthetic text. Additionally, a Turing Test confirms that our synthetic text is indistinguishable from real-world data—a result not observed with zero-shot synthetic data generation. Second, we evaluated the utility of the synthetic data by fine-tuning a BERT-sized model for text classification. Our results showed that our proposed pipeline produces higher-quality synthetic data with learning rates most similar to models fine-tuned exclusively on real-world data. Third, we demonstrate that our approach narrows the performance gap between real and synthetic data—defined as the loss in accuracy when using synthetic data instead of real-world data—on classification tasks. All experiments were conducted using the CheXpert dataset, a real-world EHR-derived collection that includes labels for concepts commonly found in chest radiology reports.

## 2 Methods

**2.1 Data Sources** This study utilized the CheXpert dataset, a large-scale open-source dataset comprising radiology reports from Stanford Hospital. CheXpert is widely recognized for its comprehensive annotations of various clinical entities. For this paper, we focused on five specific clinical entities: cardiomegaly, pneumothorax, pneumonia, pulmonary edema, and pleural effusion. These concepts were selected due to their clinical significance and the availability of annotated data within the CheXpert dataset[30].

**2.2 Dataset Preparation** For each of the five evaluation concepts, we utilized the CheXpert agent's classification labels to randomly sample from the CheXpert dataset, which contains 200,000 unique clinical notes. Specifically, we selected 100 notes indicating the presence of each concept and 100 notes indicating its negation or absence, resulting in a held-out test set of 200 unique notes for each variable. From the remaining notes, we randomly selected 5,000 clinical notes to serve as our working dataset.

**2.3 Data Annotation** To establish a reliable ground truth for the held-out test sets, we discarded the CheXpert agent's labels and had a clinician manually annotate each note for the presence or absence (or negation) of the target concept. This held-out test set was used to evaluate the performance of our fine-tuned classification models.

**2.4 Diversity Sampling Method** To generate synthetic clinical text that accurately reflects the syntactic diversity of clinical notes, we first created contextualized sentence embeddings from the clinical notes using the SFR-Embedding-Mistral model. This model was chosen for its strong performance in the Massive Text Embedding Benchmark (MTEB)[31], making it well-suited for capturing the nuanced semantics of clinical text. We applied UMAP to reduce the dimensionality of these embeddings to two dimensions. Following dimensionality reduction, we used k-means clustering to partition the data into 50 clusters. Each cluster was represented by its centroid, which served as the representative data point. These 50 representative data points, known as the diverse data, were selected to cover a broad range of the semantic space. The data points were then labeled for use in downstream few-shot synthetic data generation (Figure 1).

**2.5 Labeling and Prompt Generation** To automatically label the selected diverse data points, we employed the CLEAR pipeline[32], which leverages information retrieval upstream of LLM classification. For generating synthetic clinical notes, we used LLaMA-3.1 8B Instruct[33] (hereafter referred to as LLaMA). To optimize the quality of the generated text, we performed prompt engineering by carefully crafting prompts that would guide the model to produce realistic clinical notes, reflecting the terminology and structure found in actual clinical notes. Additionally, we incorporated few-shot prompting to enhance the synthetic data generation process. We crafted two types of prompts—one for each disease classification required by our fine-tuned BERT model. Specifically, we randomly sampled five notes indicating the presence of disease or five notes indicating the absence/negation of disease from the 50 diverse data points, repeating this process 325 times for each classification. This resulted in 650 unique prompts, each containing five-shot learning examples. Incorporating few-shot prompting helps LLaMA generate new synthetic notes that are consistent with patterns observed in real clinical data. To account for variability in text length and complexity, we calibrated the prompts to prompt LLaMA to generate notes with a token count within the 25th to 75th percentile range (114 to 192 tokens) observed in the larger population of radiology reports in the CheXpert dataset (Figure 1).

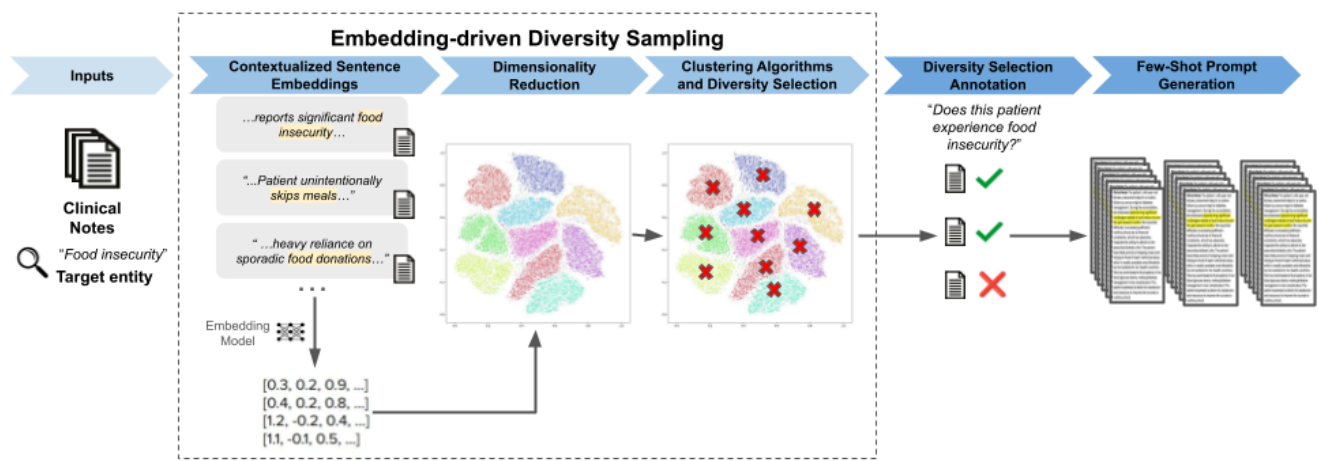

Figure 1: Diversity sampling synthetic data generation pipeline

**2.6 Control Datasets** To validate the effectiveness of our pipeline, we validated our results against three control datasets. In the first dataset, known as 'random sampling', we randomly sampled 50 notes from the working dataset for few-shot synthetic data generation. The second dataset, known as 'zero-shot sampling', leveraged a zero-shot approach to synthetic data generation, prompting LLaMA to produce synthetic notes without providing any examples of real notes and focusing solely on prompt engineering to enhance note quality. In the third dataset, known as 'real-world sampling', we randomly sampled from all notes from the working dataset. This comprehensive comparison provided insights into the effectiveness of our synthetic data generation pipeline and its ability to produce high-quality data for clinical applications.

**2.7 Synthetic Data and Real-World Augmentation Datasets** Each prompt was used to generate one synthetic data point. For each of the five variables, we generated 650 synthetic data points across all conditions—including our pipeline and the two control methods—with 325 notes indicating the presence of the concept and 325 suggesting its absence or negation. For our full real-world comparison, we randomly sampled 650 data points from the working dataset and labeled them using the CLEAR pipeline. We refer to these datasets as our augmentation datasets.

**2.8 Synthetic Data Authenticity** To assess the authenticity of our synthetic data, we performed a Turing Test involving an expert clinician. We randomly selected 100 synthetic notes and asked the clinician to distinguish them from real clinical notes. Additionally, we utilized the SFR-Embedding-Mistral model to generate contextualized sentence embeddings for the embedding-driven diverse sampling, random sampling, zero-shot synthetic data, and the real-world working dataset. We then calculated the average cosine distance between the embeddings of the synthetic datasets and those of the real-world data to quantitatively assess their similarity.

**2.9 Model Fine-Tuning and Learning Rate Analysis** We investigated whether synthetic data could be used to fine-tune a BERT model for binary classification (presence vs. negation/absence) of each of our five variables. Specifically, we fine-tuned Bio+Clinical BERT, which is initialized from BioBERT and pre-trained on all MIMIC notes[34]. For each testing condition, we defined fine-tuning datasets for each variable, starting with a baseline of 50 real-world data points in the training dataset. In the few-shot prompting experiments (both diversity sampling and random sampling), these 50 data points were the same ones used for few-shot prompting. For the zero-shot and real-world sampling, the 50 data points were randomly sampled from the working dataset. The baseline data were supplemented with data from the augmentation datasets to build the training dataset. To evaluate the learning rates for each condition, we first calculated the performance of the fine-tuned model on the held-out test set using only the baseline 50 data points. For each test condition, we then incrementally added 25 randomly selected additional data points from the corresponding augmentation dataset to the baseline data, fine-tuned the classifier, and recalculated performance metrics on the same held-out test set. We performed this process for 15 iterations, with the final iteration consisting of a model trained on 50 baseline training points and 375 augmentation training points. For fine-tuned model performance on the held-out test set, we report the mean area under the receiver operating characteristics (AUROC) and the mean area under the precision-recall curve (AUPRC) with 95% confidence intervals.

# 3 Results

**3.1 Diversity sampling produces synthetic data most similar to real-world data** Our results show the average cosine similarity distance between embedding-based retrieval of examples for few-shot prompting was most similar to real-world data when compared to zero-shot synthetic data generation methods (Table 1). This aligns with the results of the Turing Test, where an expert clinician attempted to distinguish between real and synthetic clinical notes. The results in Table 2 suggest that synthetic notes generated using the diverse and random sampling methods closely mimic real notes, making them difficult to distinguish, while zero-shot synthetic notes were more easily identified as artificial.

| Entity | Diversity Sampling | Zero-Shot Sampling | Random Sampling |
|---|---|---|---|
| Cardiomegaly | 0.81 | 0.82 | 0.72 |
| Pleural Effusion | 0.82 | 0.82 | 0.73 |
| Pneumonia | 0.83 | 0.83 | 0.74 |
| Pneumothorax | 0.83 | 0.82 | 0.74 |
| Pulmonary Edema | 0.82 | 0.83 | 0.73 |
| Average | 0.82 | 0.82 | 0.73 |

Table 1: Average cosine similarity distance between real-world data and our different synthetic data generation methods.

| Method | Correctly Identified Synthetic Notes (n, %) | Correctly Identified Real Notes (n, %) | P-value |
|---|---|---|---|
| Diverse Sampling | 26 (52%) | 28 (56%) | 0.37 |
| Random Sampling | 28 (56%) | 29 (58%) | 0.19 |
| Zero-Shot Sampling | 48 (96%) | 46 (92%) | <0.001 |

Table 2: Turing Test for three synthetic data generation methods with p-value calculated via a binomial test.

**3.2 Diversity sampling enhances synthetic data learning rates** To further illustrate the effectiveness of our diversity sampling synthetic data generation approach during the training process, Figure 2 presents a learning rate analysis tracking the AUROC performance of Bio+Clinical BERT over different model iterations where every iteration adds additional fine-tuning data from the corresponding augmentation dataset. Every model iteration was evaluated on a classification task using the held-out test set. Our results suggest models fine-tuned with the diverse sampling approach generally achieved more efficient fine-tuning than those using random and zero-shot sampling methods. This was particularly evident as the number of training steps increased and in the reduced number of steps required for the mean AUROC and mean AUPRC to reach 0.85. This analysis showed that the average steps between real-world sampling and diversity sampling is approximately 0.6 for AUROC and 1.4 for AUPRC. This represents nearly a 40% improvement in AUROC and a 30% improvement in AUPRC over the random and zero-shot sampling methods (Table 3). Models trained on real-world sampling data achieved the most efficient learning and lowest average steps to reach our 0.85 performance threshold.

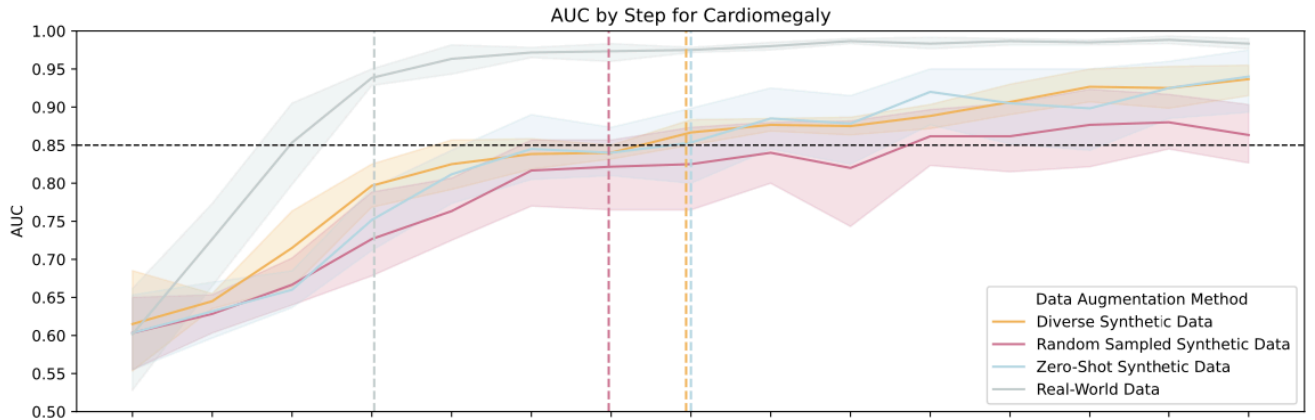
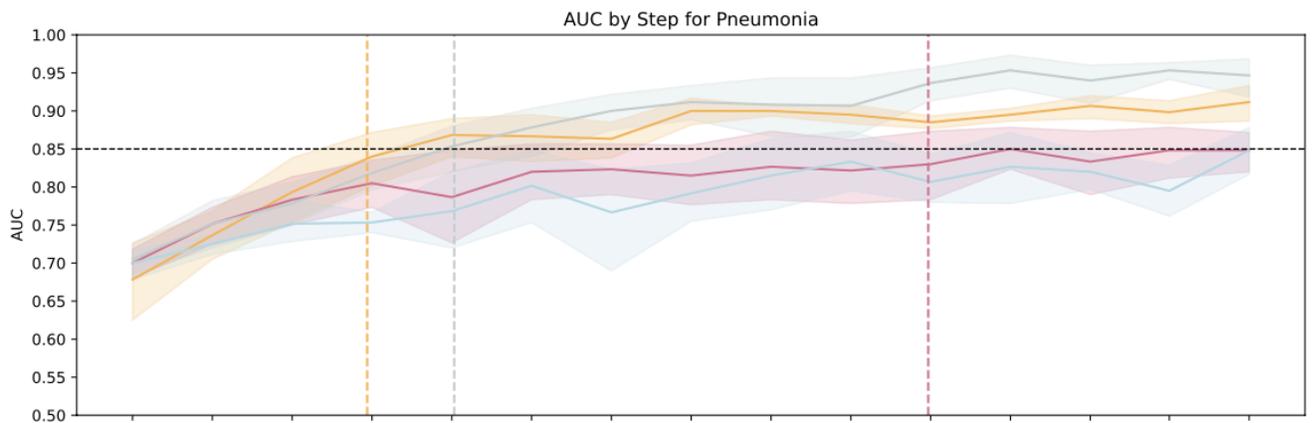
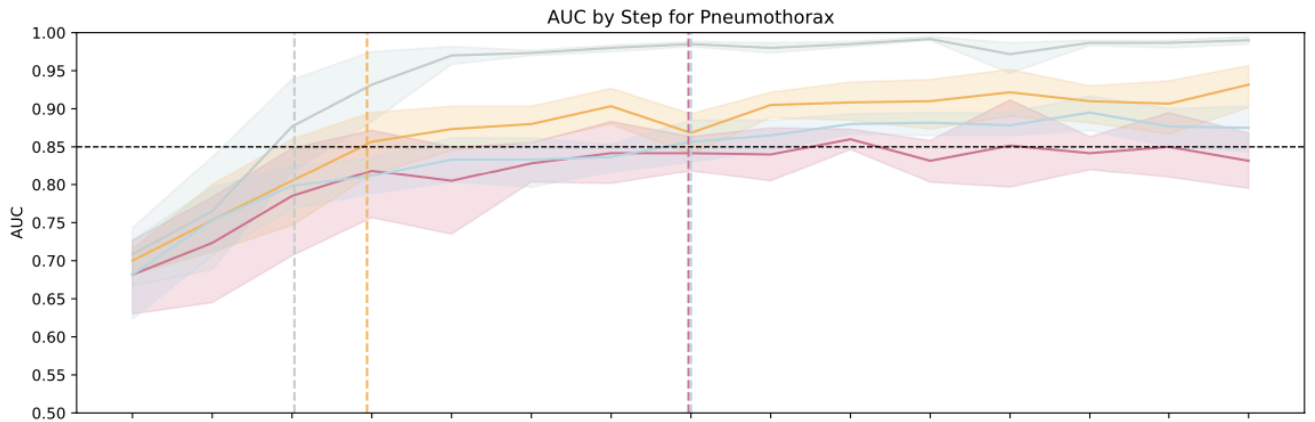
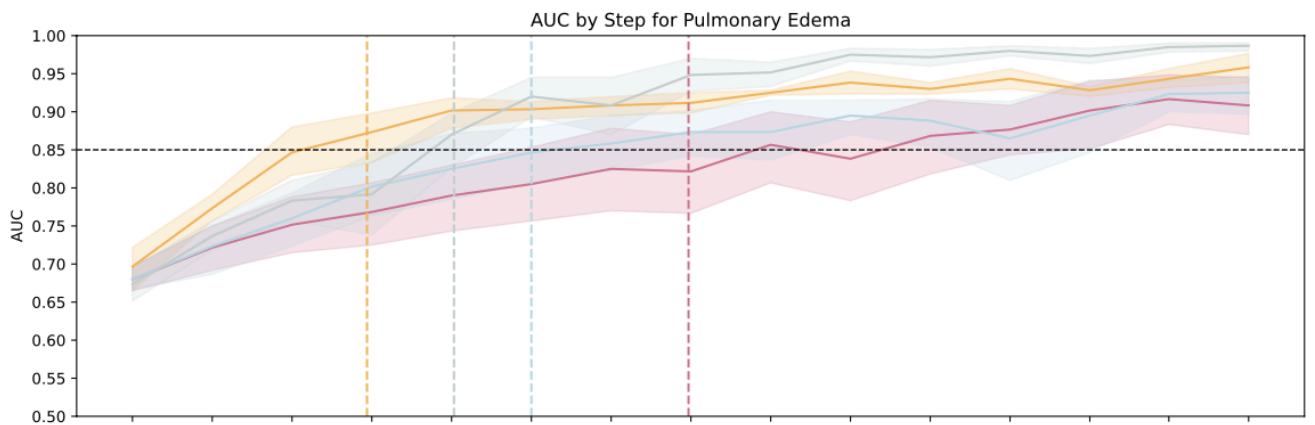
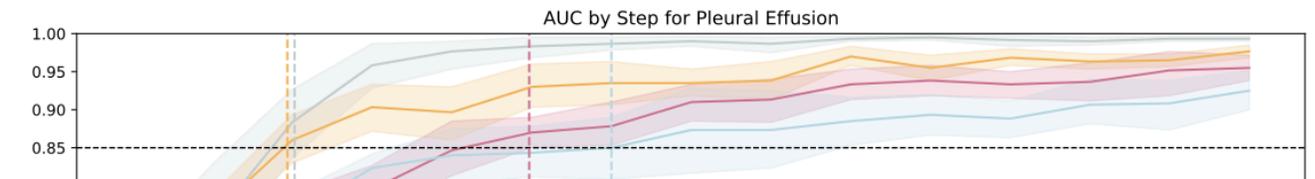

Figure 2: Learning rate analysis across different sampling methods reporting AUROC across all 15 model iterations for five clinical entities. The dashed line represents the step at which AUROC surpasses the 0.85 performance threshold.

|  | Average Steps to Reach Performance Threshold of 0.85 | |
|---|---|---|
| **Data Augmentation Method** | **AUROC** | **AUPRC** |
| Real-World Sampling | 4.0 | 4.8 |
| Diversity Sampling | 4.6 | 6.2 |
| Random Sampling | 8.0 | 8.5 |
| Zero-shot Sampling | 7.2 | 9.0 |

Table 3: Average steps required for classification models to reach 0.85 AUROC and AUPRC performance thresholds.

**3.3 Diversity sampling closes the real-to-synthetic data gap** For all models, the augmentation datasets improved both AUROC and AUPRC, with diversity sampling showing the greatest improvement over random and zero-shot sampling (Figure 3). Our findings reveal that the average difference between real-world sampling and diversity sampling is 4.1% for both metrics. This translates to approximately a 57% improvement in AUROC and a 68% improvement in AUPRC compared to random and zero-shot sampling methods (Table 4). We also calculated the total number of augmentation data point required for the classification models to reach an AUROC of 0.85. By comparing the ratio of real to synthetic data, we found that one diversity sampled data point is approximately 0.9 times as effective as a real-world data point which is nearly a 60% improvement in worth compared to random and zero-shot sampling methods (Table 5).

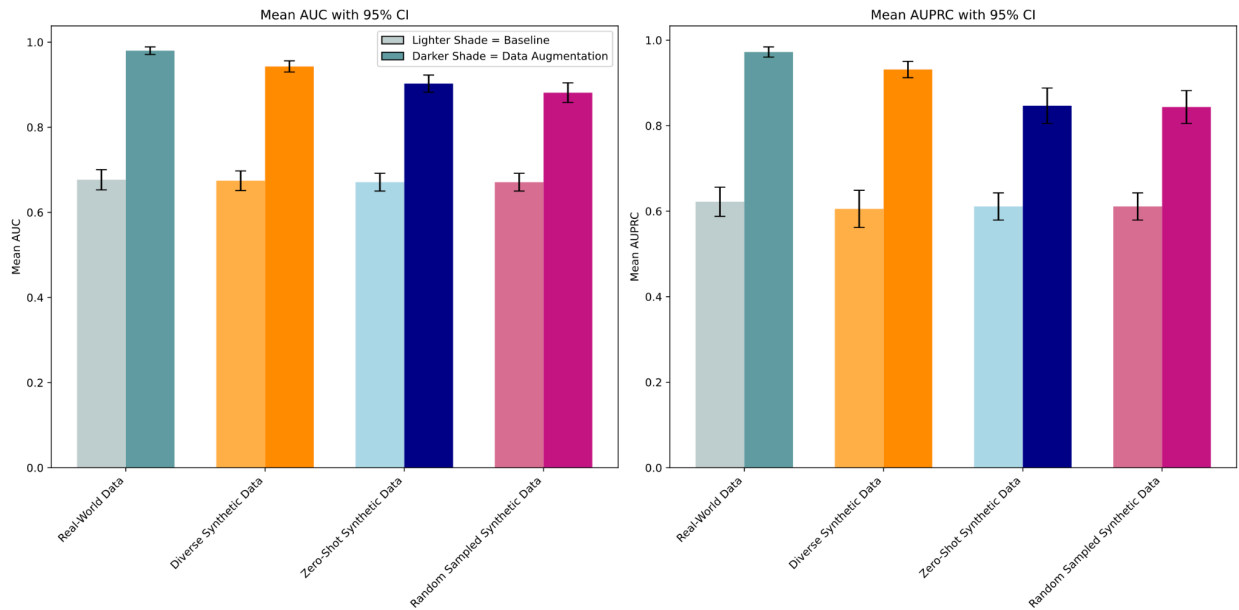

Figure 3: Barplot of baseline and final data augmentation results averaged across all five entities. Baseline models include 50 data points with the augmentation dataset adding 375 additional data points.

| Data Augmentation Method | Final Augmented Model [CI: 95%] | | Performance Gap Between Real-World and Synthetic Data Augmentation (%) | |
|---|---|---|---|---|
| | AUROC | AUPRC | AUROC | AUPRC |
| Real-World Sampling | 0.98 [0.97-0.98] | 0.97 [0.96-0.98] | | |
| Diversity Sampling | 0.94 [0.93-0.96] | 0.93 [0.91-0.95] | 4.1% | 4.1% |
| Random Sampling | 0.88 [0.85-0.90] | 0.84 [0.81-0.88] | 10.6% | 13.2% |
| Zero-shot Sampling | 0.90 [0.88-0.92] | 0.85 [0.81-0.89] | 8.5% | 12.9% |

Table 4: Data augmentation model performance gap compared to real-world model.

| Data Augmentation Method | Additional Data to Reach AUROC Threshold of 0.85 | Real-to-Synthetic Data Ratio |
|---|---|---|
| Real-World Sampling | 100 | |
| Diversity Sampling | 112 | 0.90 |
| Random Sampling | 177 | 0.56 |
| Zero-shot Sampling | 176 | 0.57 |

Table 5: Total augmentation data points required to reach AUROC threshold of 0.85 with real-to-synthetic data ratio.

## 4 Discussion

In this study, we introduced an embedding-driven diversity sampling approach for synthetic data generation in clinical NLP tasks. By leveraging contextualized sentence embeddings, our diversity sampling method provided LLMs with a broad spectrum of examples that captured the syntactic diversity of real-world clinical text. Our findings highlight the potential of this approach to enhance the quality of synthetic clinical notes, resulting in notable improvements in both the efficiency and performance of BERT-sized models on clinical classification tasks.

The findings from our experiments indicate that our embedding-driven diversity sampling synthetic data approach, referred to as diversity sampling, outperformed both random and zero-shot sampling methods in several key aspects. The synthetic data produced using diversity sampling closely mirrored the style and quality of real-world clinical notes which resulted to better fine-tuned model performance. Specifically, models trained with diversity sampling were able to reduce the real-to-synthetic data performance gap. Furthermore, we show that our diversity-sampled data points more closely match the quality of real-world data compared to random and zero-shot sampling, falling just 10% short of full equivalence.

Our results highlight the strengths of the embedding-driven diversity sampling approach over previous synthetic data generation methods. Traditional approaches like GatorTronGPT, while effective, suffer from limitations in generalizability across different healthcare systems and are computationally expensive to develop. Other studies, such as those using random or zero-shot sampling for few-shot learning, fail to capture the syntactic diversity and complexity of real-world clinical language.

The promising performance of the diversity sampling approach has several important implications for clinical NLP applications. First, it provides a scalable solution to the challenge of limited annotated data, offering a means to bootstrap synthetic data when access to real-world annotated clinical notes is restricted. The enhanced diversity captured by our method can be instrumental in improving the accuracy of models designed for tasks such as electronic phenotyping, cohort selection, and predictive modeling.

Furthermore, by bridging the performance gap between synthetic and real-world data, our approach can aid in building more robust and generalizable NLP models that perform well across different healthcare settings.
Despite the strengths of our approach, there are some limitations to consider. This study focused on a specific dataset (CheXpert) and five targeted clinical entities, which may limit the generalizability of the results to other types of clinical notes or domains. Further validation on a broader set of clinical NLP tasks is necessary to fully establish the robustness of this method. Additionally, while our approach is more computationally feasible compared to some previous models, generating and validating synthetic notes still requires a non-trivial amount of computational resources and manual effort for evaluation.

To further expand on this study, future research could explore the utility of embedding-driven synthetic data generation in other clinical domains and datasets. Moreover, integrating this method with other synthetic data generation techniques, such as variational autoencoders or advanced generative models, could potentially enhance the quality and utility of the synthetic data produced. Finally, more extensive validation involving clinician assessments could provide deeper insights into the practical applicability of embedding-driven synthetic data in real-world clinical NLP tasks.

**5 Conclusion**

In conclusion, this study demonstrates the potential of embedding-driven combined with diversity sampling synthetic data generation to enhance clinical NLP applications. By accurately capturing the syntactic diversity of real-world clinical notes, our approach significantly improves model performance and efficiency, offering a viable solution for scenarios where aggregating expert annotated data is limited. These findings suggest that embedding-driven diversity sampling synthetic data can serve as an effective supplement to real-world data in developing clinical text classification models.